\def\year{2020}\relax
\documentclass[letterpaper]{article} 
\usepackage{aaai20}  
\usepackage{times}  
\usepackage{helvet} 
\usepackage{courier}  
\usepackage[hyphens]{url}  
\usepackage{graphicx} 
\usepackage{graphicx}  
\usepackage{latexsym}
\usepackage{amsmath}
\usepackage{amsfonts}
\usepackage{multirow}
\usepackage{caption}
\usepackage{color}
\usepackage[textsize=scriptsize]{todonotes}
\definecolor{blue}{RGB}{0, 93, 170}			
\definecolor{darkgreen}{RGB}{0, 102, 0}
\newcommand{\citet}[1]{\citeauthor{#1}~\shortcite{#1}}
\newcommand{\citep}{\cite}

\title{Infusing Knowledge into the Textual Entailment Task\\ Using Graph Convolutional Networks}

\author{
 Pavan Kapanipathi$^\dagger$, Veronika Thost$^\ast$$^\dagger$, Siva Sankalp Patel$^\dagger$, Spencer Whitehead$^\S$, \\ 
\Large \bf Ibrahim Abdelaziz$^\dagger$, Avinash Balakrishnan$^\dagger$, Maria Chang$^\dagger$, Kshitij Fadnis$^\dagger$, Chulaka \\ \Large \bf Gunasekara$^\dagger$, Bassem Makni$^\dagger$, Nicholas Mattei$^\ddagger$, Kartik Talamadupula$^\dagger$, Achille Fokoue$^\dagger$
\AND
{\normalsize \rm $^\dagger$ IBM Research, $^\ast$ MIT-IBM Watson AI Lab, $^\S$ University of Illinois at Urbana-Champaign, $^\ddagger$ Tulane University}\\
\small{\{kapanipa, avinash.bala, kpfadnis, krtalamad, achille\}@us.ibm.com} \\ \small{\{veronika.thost, siva.sankalp.patel, ibrahim.abdelaziz1, maria.chang, chulaka.gunasekara, bassem.makni\}@ibm.com}\\
\small{srw5@illinois.edu}\\
\small{nsmattei@tulane.edu}\\
}

\begin{document}
\maketitle
\begin{abstract}
Textual entailment is a fundamental task in natural language processing. 
Most approaches for solving this problem use only the textual content present in training data. 
A few approaches have shown that information from external knowledge sources like knowledge graphs (KGs) can add value, in addition to the textual content, by providing background knowledge that may be critical for a task. However, the proposed models do not fully exploit the information in the usually large and noisy KGs, and it is not clear how it can be effectively encoded to be useful for entailment.
We present an approach that complements text-based entailment models with information from KGs by (1)~using Personalized PageRank to generate contextual subgraphs with reduced noise and (2)~encoding these subgraphs using graph convolutional networks to capture the structural and semantic information in KGs.
We evaluate our approach on multiple textual entailment datasets and show that the use of external knowledge helps the model to be robust and improves prediction accuracy. This is particularly evident in the challenging BreakingNLI dataset, where we see an absolute improvement of 5-20\% over multiple text-based entailment models.
\end{abstract}

\section{Introduction}
Given two natural language sentences, a premise \textsc{P} and a hypothesis \textsc{H}, the textual entailment task -- also known as natural language inference (NLI) -- consists of determining whether the premise entails, contradicts, or is neutral with respect to the given hypothesis~\cite{maccartney2009}.  In practice, this means that textual entailment is characterized as either a three-class (\textsc{entails/neutral/contradicts}) or a two-class (\textsc{entails/neutral}) classification problem~\cite{bowman2015large,khot2018scitail}. 

Performance on the textual entailment task can be an indicator of whether a system, and the models it uses, are able to reason over text. This has tremendous value for modeling the complexities of human-level natural language understanding, and in aiding systems tuned for downstream tasks such as question answering~\cite{harabagiu2006methods}.

\begin{figure}[htp]
  \centering
    \includegraphics[width=0.5\textwidth]{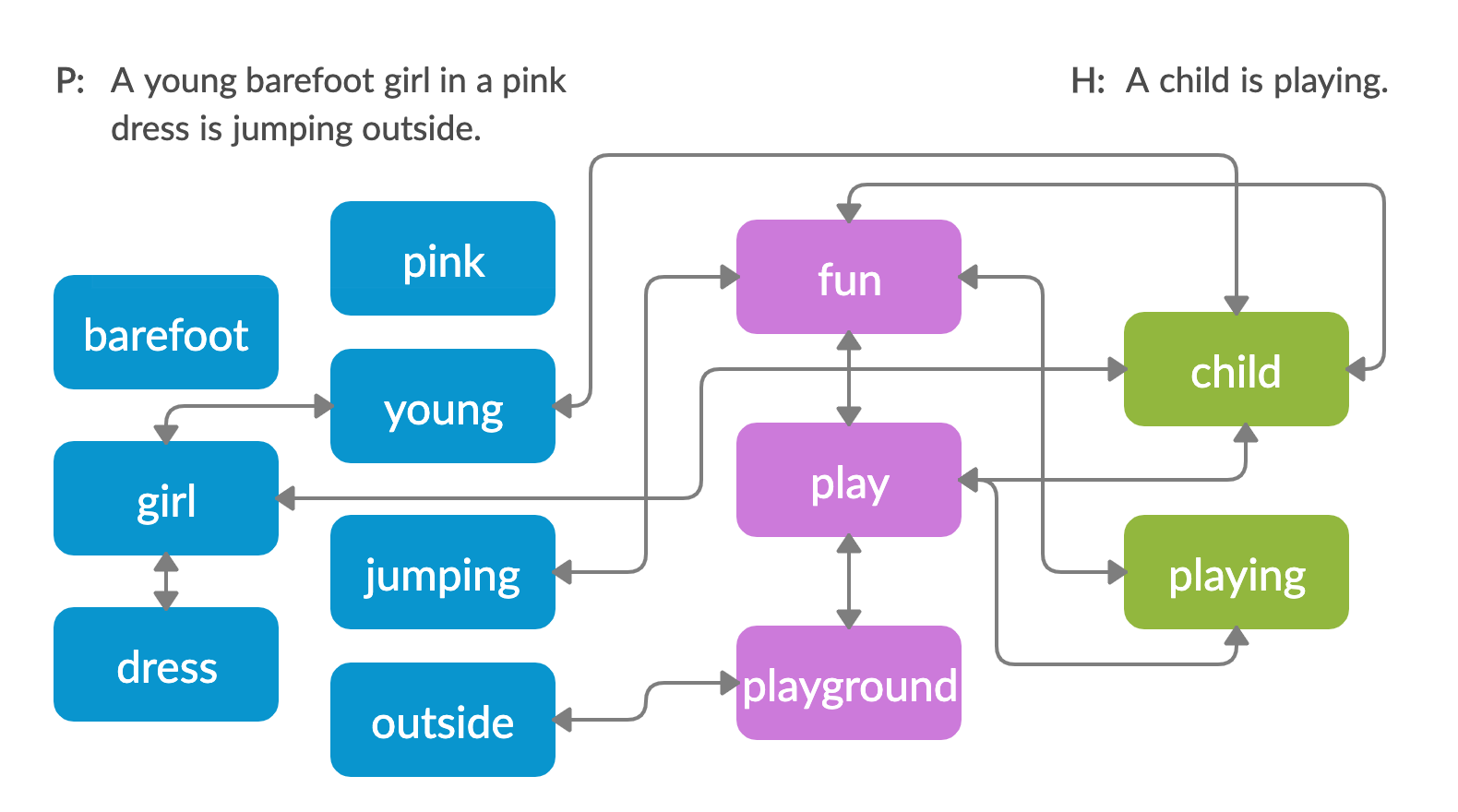}
    \caption{A premise and hypothesis pair along with a relevant subgraph from ConceptNet. Blue concepts occur in the premise, green in the hypothesis, and purple connect them.} 
    \label{fig:kg-example}
\end{figure}

Most existing textual entailment models focus only on the text of premise and hypothesis to improve classification accuracy~\cite{parikh2016decomposable,liu2019multi}. A recent and promising line of work has turned towards extracting and harnessing relevant semantic information from knowledge graphs (KGs) for each textual entailment pair~\cite{chen2018neural,wang2019improving}. These approaches map terms in the premise and hypothesis text to concepts in a KG, such as Wordnet~\cite{miller1995wordnet}, ConceptNet~\cite{speer2017conceptnet}, or DBpedia~\cite{auer2007dbpedia} and use these mapped concepts for the textual entailment task. Figure~\ref{fig:kg-example} shows an example of such mapping, where select terms from the premise and hypothesis are mapped to concepts from a knowledge graph (blue and green nodes, respectively). However these models suffer from one or more of the following drawbacks: (1) they do not possess the ability to explicitly capture the semantic and structural information from the KG. For example, in Figure~\ref{fig:kg-example}, the ability for models to encode information from paths between blue and green nodes via purple nodes provides better context facilitating the system to more correctly judge entailment.; (2) they are not easily integrated with existing NLI models that exploit only the text of the premise and hypothesis; and (3) they are not flexible with respect to the type of KG that is used. 

\smallskip
\noindent
\textbf{Contributions}: We present an approach to the NLI problem that can augment any existing text-based entailment model with external knowledge. We specifically address the aforementioned challenges by: (1)~introducing a neighbor-based expansion strategy {in combination with} subgraph filtering using Personalized PageRank (PPR) \cite{jeh2003scaling}. This approach reduces noise and selects contextually relevant subgraphs from larger external knowledge sources for premise and hypothesis texts ; (2)~encoding subgraphs using Graph Convolutional Networks (GCNs) \cite{kipf2017gcn}, which are initialized with knowledge graph embeddings to capture structural and semantic information. This general approach to graph encoding allows us to use any external knowledge source that can be represented as a graph such as WordNet, ConceptNet, or DBpedia.  We show that the additional knowledge can improve textual entailment performance by using four standard benchmarks: SciTail, SNLI, MultiNLI, and BreakingNLI. In particular, our experiments on the BreakingNLI dataset, where we see an absolute improvement of 3-20\% over four text-based models, shows that our technique is robust and resilient.

\section{Related Work}
\label{sec:related_work}
We categorize the related approaches for NLI into: (1) approaches that take only the premise and hypothesis text as input, and (2) approaches that utilize external knowledge. 

Neural models focusing solely on the textual information \cite{wang2016learning,yang2019enhancing} explore the sentence representations of premise 
structure and max pooling layers. Match-LSTM \cite{wang2016learning} and Decomposable Attention \cite{parikh2016decomposable} learn cross-sentence correlations using attention mechanisms, where the former uses a asymmetric network structure to learn premise-attended representation of the hypothesis, and the latter a symmetric attention, to decompose the problem into sub-problems. Latest NLI models  use tranformer architectures such as BERT \cite{devlin2019bert} and RoBERTa \cite{liu2019roberta}. These models perform exceedingly well on many NLI leaderboards \cite{zhang2018know,liu2019multi}. In this work, we show that performance of text-based entailment models that use pre-trained BERT embeddings can be augmented with external knowledge. 

Utilizing external knowledge has shown improvement in performance on many natural language processing (NLP) tasks~\cite{huang2019knowledge,moon2019opendialkg,musa2019answering}. Recently, for NLI,  \citet{li2019several} have shown that features from pre-trained language models and external knowledge  complement each other.  However, approaches that do utilize external knowledge for NLI are very few~\cite{wang2019improving,chen2018neural}. In particular, the best model of \citet{wang2019improving} combines rudimentary node information -- in the form of concepts mentioned in premise and hypothesis text (blue and green nodes in Figure~\ref{fig:kg-example}) -- along with the text information. However, this approach misses the rich subgraph structure that connects  premise and hypothesis entities (purple nodes in Figure~\ref{fig:kg-example}). 
\cite{chen2018neural}~have developed a model with WordNet based co-attention that use five engineered features from WordNet for each pair of words from premise and hypothesis. This model being tightly integrated with WordNet has the following drawbacks: (1) it is inflexible to be used with other external knowledge sources such as ConceptNet or DBpedia, and (2) it is non-trivial to be integrated with other state of the art text-based entailment systems. This work addresses the drawbacks of each of these approaches mentioned above with competitive performance on many NLI datasets.

The availability of large-scale datasets \cite{bowman2015large,williams2018broad,khot2018scitail} has fostered the advancement of neural NLI models in recent years. However, it is important to discuss the characteristics of these datasets to understand what they intend to evaluate~\cite{glockner2018breaking}. Particularly, datasets such as \cite{bowman2015large,khot2018scitail,williams2018broad} contain language artifacts as significant cues for text-based neural models. These artifacts bias the models and makes it harder to evaluate the impact of external knowledge \cite{chen2018neural,wang2019improving}. In order to evaluate approaches that are more robust and not susceptible to such biases, \citet{glockner2018breaking} created BreakingNLI -- an adversarial test set where most of the common text-based approaches show significant drop in performance. It is  important to note that this test set is generated using a subset of relationships from online resources for English learning, making it more suitable for models exploiting KGs with lexical focus, such as WordNet. However, BreakingNLI represents a first and important step in the evaluation of models that utilize external knowledge sources. 

One of the core contributions of this work is the application of Graph Convulutional Networks for encoding knowledge graphs. While (graph-)structured knowledge represents a significant challenge for classical machine learning models, 
graph convolutional networks \cite{kipf2017gcn} offer an effective framework for representation learning of graphs. In parallel, relational GCNs (R-GCNs) \cite{schlichtkrull2018modeling} have been designed to accommodate the highly multi-relational data characteristics of large KGs. Inspired by these works, we explore the use of R-GCNs for infusing information from KG into NLI models.

\section{KG-Augmented Entailment Model}
\label{sec:approach}

\begin{figure}[b!]%
  \centering%
    \includegraphics[width=0.4\textwidth]{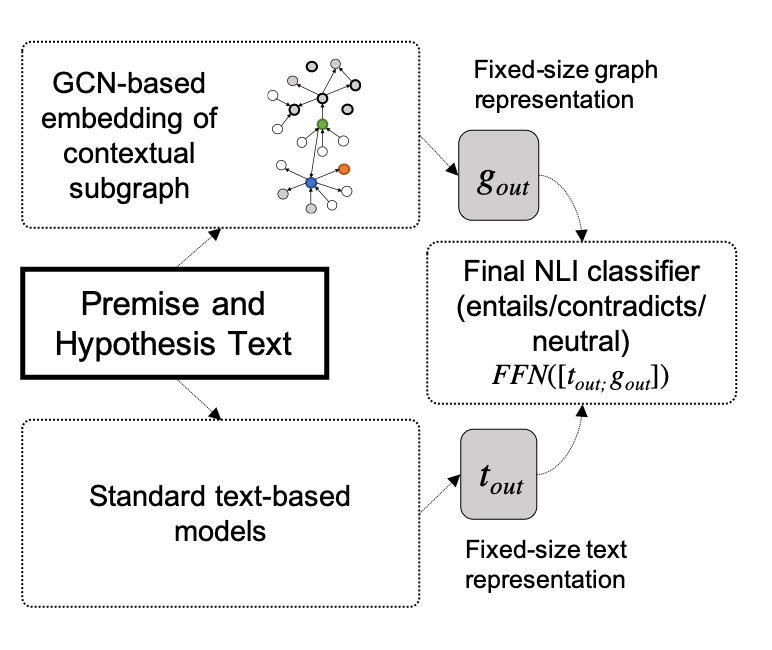}
      \caption{
      Primary components of KES: standard text-based model, GCN-based graph embedder, and final feedforward classifier.
      }
    \label{fig:overview_figure}
\end{figure}

In this section, we describe the central contribution of the paper -- the KG-augmented Entailment System (KES). As shown in Figure~\ref{fig:overview_figure}, KES consists of two main components.  
 The first component is a standard text-based model that creates a fixed-size representation of the premise and hypothesis texts. 
The second component selects contextual subgraphs for the premise and the hypothesis from a given KG, and encodes them using a GCN.  
The fixed size representations from the two components are used as input to a standard feedforward layer for classification.
We opted for a combined graph and text approach because the noise and incompleteness of KGs renders a purely graph-based approach insufficient as a standalone solution.
However, we 
show that the KG-augmented model provides valuable context and additional knowledge that may be missing in text-only representations. 

\begin{figure*}
  \centering
    \includegraphics[width=1\textwidth]{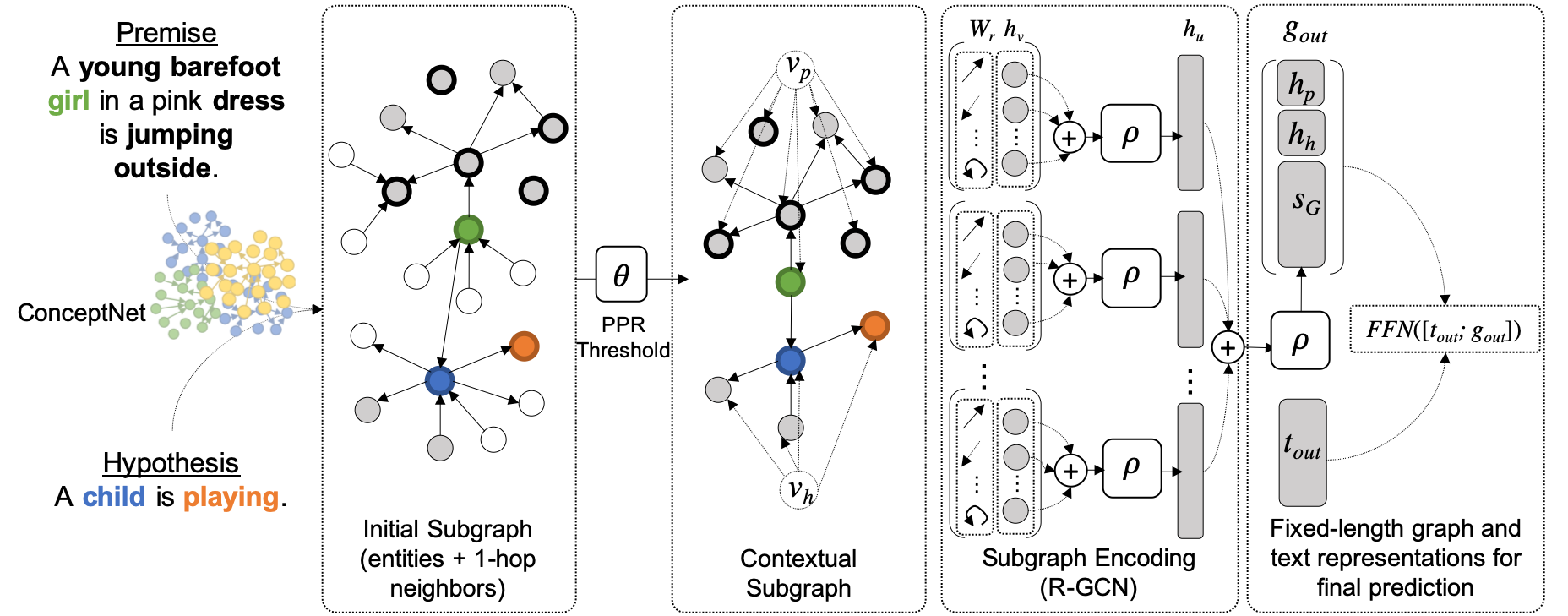}

    \caption{
    Overview of the KES approach. KES links terms in the premise and hypothesis to concepts in KG, creates contextual subgraphs via personalized page rank filtering, encodes those subgraphs with an R-GCN, and finally combines the aggregated node embeddings with text representations into a feedforward classifier.
    $h_p$ and $h_h$ in the figure denote $h^{L}_{v_p}$ and $h^{L}_{v_h}$ in  Equation~\eqref{eq:gcn_encoding} respectively.
    }
    \label{fig:graph_approach}
\end{figure*}

\subsection{A Standard Text-based Model} 
Given the premise $P = (p_1, \dots, p_n)$ and hypothesis $H = (h_1, \dots,  h_m)$, let $p_i$ and $h_j$ be the embeddings of words occurring in sequence in the premise and hypothesis texts. These embeddings are input to a neural network $T_{\textit{NLI}}$ that outputs a fixed size representation $t_{out}\in\mathbb{R}^{K}$ :

\begin{align}
t_{out} &= T_{\textit{NLI}}(P, H)
\label{eq:text_model}
\end{align}
 where $T_{\textit{NLI}}$ can be any of the existing state of the art text-based NLI models \cite{wang2016learning,talman2019sentence,liu2019multi}.

\subsection{Contextual Subgraphs and their Representation using GCNs}
This component uses an external KG to obtain a subgraph that is relevant with respect to the premise and the hypothesis, and then applies  GCN to encode this subgraph into a fixed-size representation $g_{out}$ (Figure~\ref{fig:graph_approach}).  

\subsubsection{Subgraph Extraction:}
In order to retrieve a subgraph from the KG, we first map the terms in premise and hypothesis text to concepts in KG by performing a max-substring match. For example, given the premise and hypothesis in Figure~\ref{fig:kg-example}, the extracted and mapped concepts are shown in blue and green. Next, this initial set of concepts is then expanded to include (one-hop) neighbor concepts,  and all the edges between them (initial set and their neighbors) from the KG. In the example in Figure~\ref{fig:kg-example}, we extract a subgraph that includes  the purple nodes because they are directly connected to green and/or blue nodes.

\subsubsection{Personalized PageRank (PPR) to Filter Context:} 
KGs are typically very large, and concept expansion by just one hop can introduce a significant amount of noise \cite{wang2019improving,lalithsena2017domain}. For example, the concept \texttt{girl} is directly connected to over 1000 other concepts in ConceptNet.  
For this reason, we create a {\em contextual} subgraph by further filtering the one-hop subgraph.

To obtain the most relevant neighbor nodes given the premise and hypothesis texts, we use Personalized PageRank (PPR) \cite{page1999pagerank}. 
PPR adds a bias to the PageRank algorithm by scoring the nodes conditioned on a initial subset of nodes in the graph. The bias is introduced by changing the uniformly distributed jump probability vector \textbf{p} of PageRank to a non-uniform distribution with respect to the initial subset of nodes (Equation~\ref{eq:jump_probablitiy_vector}). In our settings, this initial subset of nodes $S$  consists of the concepts mentioned in the premise and hypothesis.  
\begin{equation}
p_i = \begin{cases}
    \dfrac{1}{|S|}       & \quad i \in S\\
    \ \ 0  & \quad i \notin S
  \end{cases}
\label{eq:jump_probablitiy_vector}
\end{equation}
PPR-scores $\mathbf{R'}$ are then computed as follows: 
\begin{equation}
\mathbf{R'}= (1 - \alpha) \mathbf{A} \times \mathbf{R} + \alpha \mathbf{p}
\label{eq:pagerank}
\end{equation}
where $\mathbf{R}$ is a vector with scores for each node (post convergence); $\mathbf{A}$ is a normalized adjacency matrix (transition probability matrix); and $\alpha$ is the damping factor. 

We normalize the PPR-scores based on the maximum PPR-score of a node in the sub-graph. We then choose a filtering threshold $\theta$, and exclude all the nodes that are not in the initial subset $S$ and that have a PPR-score below $\theta$; we also exclude the edges that link to the deleted nodes. The remaining nodes and edges make up the contextual subgraph for the premise-hypothesis pair under consideration.

\subsubsection{Encoding Contextual Subgraphs:}
The contextual subgraph for premise and hypothesis is encoded using a relational graph convolutional network (R-GCN)~\cite{schlichtkrull2018modeling}. 
GCNs compute node embeddings by iteratively aggregating the embeddings of neighbor nodes.
R-GCNs extend standard GCNs ~\cite{kipf2017gcn} to deal with the multi-relational data of KGs. They learn different weight matrices for each type of relation occurring in the graph. We use an R-GCN to compute node embeddings, and then aggregate these embeddings to obtain a fixed-size representation for the contextual subgraph. 

We first extend the contextual subgraph by adding a self-loop edge for each node; this is to retain the information of the node during convolution. Previous work~\cite{wang2019improving} showed that the concepts mentioned in premise and hypothesis played an important role to improve NLI performance. Inspired by this, we retain information of concepts (nodes) that occur in  premise  and hypothesis text by adding a premise supernode $v_p$ and hypothesis supernode $v_h$. The premise supernode is connected to concepts that are mentioned in premise using bi-directional edges and similarly the hypothesis supernode is connected to the concepts mentioned in the hypothesis.

We then apply the algorithm suggested by \citet{nguyen2018graph} -- which uses a simple sum as the aggregation function -- but we include a normalization factor and disregard bias (similar to \citet{schlichtkrull2018modeling}): 

\begin{align}
h_{u}^{l+1} &= \rho\left(  
\sum_{r\in \mathcal{R}}
\sum_{v\in\mathcal{N}_{u,r}} \frac{1}{c_{u,r}}W_{r}^{l} h_{v}^{l} \right).
\label{eq:rgcn}
\end{align}
Here, $\mathcal{R}$ is the set of edge types; 
$\mathcal{N}_{u,r}$ is the set of neighbors connected to node $u$ through the edge type $r$; $c_{u,r}$ is a normalization constant;
$W_{r}^{l}$ are the learnable weight matrices, one per edge type $r\in \mathcal{R}$; 
and $\rho$ is a non-linear activation function. 
We use the (symmetric) normalized Laplacian as a normalization constant \cite{kipf2017gcn}.

The final node embeddings are aggregated using a summation-based graph-level readout function~\cite{xu2019gnnpower}:
\begin{align}
s_{G} &= 
\rho\left(  \sum_{v\in V} W h_{v}^{L} \right).
\label{eq:aggregation_function}
\end{align}
$V$ is the set of nodes in our contextual graph, $W$ is a learnable weight matrix, and $\rho$ is an activation function. 
This summation-based readout function allows the encoder to learn representations that encode the structure of the graph.

The final representation of the contextual subgraph is obtained by concatenating  $s_{G}$ -- the aggregated embeddings of all the nodes -- 
with the embeddings of the premise and hypothesis supernodes as follows:

\begin{equation}
g_{out} = [{s}_G; h^{L}_{v_p}; h^{L}_{v_h}] \label{eq:gcn_encoding}
\end{equation}

\subsection{Final Classifier}
The final feedforward classifier takes as input the text encoding from Equation~\eqref{eq:text_model} and the graph encoding from Equation~\eqref{eq:gcn_encoding} to  classify the premise and hypothesis as entailment/contradiction/neutral: 

\begin{equation}
\mathbf{E_{pred}} = \textit{FFN}\left([t_{{out}}; g_{{out}}]\right)
\label{eq:final_feed_forward}
\end{equation}

\section{Experiments \& Results}
\label{sec:results}
In this section, we describe the experiments that we performed to evaluate our approach; the setup, including datasets, models, and implementations; and the results. 

\subsection{Datasets}
\label{subsec:experiment_datasets}
We considered the most popular NLI datasets: SNLI~\cite{bowman2015large}, SciTail~\cite{khot2018scitail},  and MultiNLI~\cite{williams2018broad}. While SNLI and MultiNLI are prominent datasets covering a wide range of topics, SciTail offers an in-depth focus on science domain questions. Since this difference is also reflected in linguistic variations, the two datasets allow evaluating very different settings. As mentioned in Section~\ref{sec:related_work}, these datasets carry linguistic cues that are easily captured by the neural text based models. Hence, to show the impact of knowledge graphs, we primarily evaluate our approach on the and BreakingNLI dataset~\cite{glockner2018breaking}.

\subsection{Knowledge Graphs}
Prior work on NLI has shown that ConceptNet contains information more useful to this problem compared to DBpedia and WordNet \cite{wang2019improving}. Furthermore, \citet{speer2017conceptnet} showed that, when ConceptNet is combined with embeddings acquired from distributional semantics, it provides applications with a richer understanding than narrower resources like the latter KGs. 
We therefore focus on ConceptNet for now, leaving experiments with other KGs as future work.

\subsection{Models for Text Representations}\label{sec:text_based_models}
We experimented with four different text-based models to obtain numerical representations of premise and hypothesis text (Equation~\eqref{eq:text_model}). Our selection criteria: (1) performance on leaderboards, (2) relevance for NLP in general, and (3) ease of implementation and availability. Our goal is to augment each of these models with external knowledge and hence test the generalizability of KES, which also shows the benefits of its modularity. We used the AllenNLP library\footnote{\url{https://github.com/allenai/allennlp}. AllenNLP includes the Decompattn model.} to implement the models described below (see also Section~\ref{sec:related_work}).%

\noindent
 \textbf{Decomposable Attention Model (DecompAttn).}
One of the earlier and most common baseline models used for NLI~\cite{parikh2016decomposable,wang2019improving,glockner2018breaking,chen2018neural}. Hence, our hypothesis is that KES can add more value and have a larger delta in performance.

\noindent
\textbf{match-LSTM.}
A NLI model with good performance on not only on multiple NLI leaderboards such as SciTail and SNLI but also applicable to other NLP tasks such as question answering~\cite{wang2016machine}. 

\noindent
\textbf{BERT + match-LSTM.} Version of match-LSTM using BERT embeddings instead of the GLoVe embeddings in the former. We opted for this model to take advantage of the improvements BERT embeddings have generated for numerous NLP tasks.

\noindent
\textbf{Hierarchical BiLSTM Max Pooling (HBMP).}
Shows superior performance on multiple NLI benchmarks including SciTail, SNLI, and MultiNLI.

\subsection{Models using External Knowledge}
There are two other models exploiting external knowledge for NLI. We compare them to KES:

\noindent
 \textbf{KIM} ~\cite{chen2018neural} uses five different features for every pair of terms from premise and hypothesis. The features are extracted from WordNet and they are infused in the model as knowledge-based co-attention mechanism. 

\noindent
\textbf{ConSeqNet} ~\cite{wang2019improving} takes the concepts mentioned in premise and hypothesis as input to a match-LSTM model (with a GRU encoding). It is important to note that the match-LSTM model better suits text than graph structure because it uses a seq2seq encoder to account for the inherent sequential nature of text, which is not present in graphs.

\begin{table*}[t!]
\resizebox{\linewidth}{!}{
\begin{tabular}{|l|c|c|c|c|c|c|c|c|}  
\hline
\multirow{2}{*}{\textbf{Models}} & \multicolumn{2}{|c|}{\textbf{Scitail}} & \multicolumn{2}{|c|}{\textbf{MultiNLI}} &
\multicolumn{2}{|c|}{\textbf{SNLI}} &
\multicolumn{2}{|c|}{\textbf{BreakingNLI}}\\

&\textbf{Text}&\textbf{KES}
&\textbf{Text}& \textbf{KES}
&\textbf{Text}&\textbf{KES}
&\textbf{Text}&\textbf{KES}\\
\hline
match-LSTM	& 82.54 & 82.22 (0.6) & 71.32 & 71.67 (0.8) & 83.60 & 83.94 (0.6) & 65.11 & \textbf{78.72} \\
BERT+match-LSTM	& 89.13 & \textbf{90.68} (0.2) & 77.96 & 76.73 (0.6) & 85.78 & 85.97 (0.6)& 59.42 & \textbf{77.59}  \\
HBMP & 81.37 & \textbf{83.49} (0.2) & 69.27 & 68.42 (0.6) & 84.61 & 83.84 (0.2) & 60.31 & \textbf{63.60} \\
DecompAttn & 76.57 & 72.43 (0.8) &  64.89 & \textbf{71.93} (0.6) & 79.28 & \textbf{85.56} (0.6) & 51.3* & \textbf{59.83} \\
\hline
\textbf{Existing Models with KG}&\textbf{Text}&\textbf{Text+Graph}
&\textbf{Text}& \textbf{Text+Graph}
&\textbf{Text}&\textbf{Text+Graph}
&\textbf{Text}&\textbf{Text+Graph}\\
\hline
KIM~\cite{chen2018neural} & - & \textbf{NE} & - & 76.4* & - & 88.6* & - & 83.1*\\
ConSeqNet~\cite{wang2019improving} & 84.2* & 85.2* & 71.32 & 70.9 &  83.60 & 83.34 & 65.11 & 61.12 \\
\hline
\end{tabular}
}
\caption{Entailment accuracy results of KES with different text models compared to text-only entailment models (Text). Bold values indicate where KES improves performance. PPR $\theta$-values are shown in parentheses.$^*$Reported values from related work.
}
\label{tab:text_graph_models_results}
\end{table*}

\subsection{Experimental Setup and Implementation}

To evaluate the impact of KES on NLI in general and its compatibility with various existing models, we compared all text-based models described above (Section~\ref{sec:text_based_models}) to a combined text+graph model.  Because the BreakingNLI test set is derived from the SNLI training set, all models trained on SNLI were evaluated on both the SNLI and BreakingNLI test sets. 

\subsubsection{Text Model Parameters.}
We chose hyperparameters as reported in related works. For match-LSTM and  BERT-match-LSTM, we refer to~\cite{wang2019improving}. For HBMP and DecomAttn, we used the parameters from \cite{talman2019sentence} and \cite{parikh2016decomposable} respectively.  

\subsubsection{KES Setup and Training.}
As initial graph embeddings, we considered TransH~\cite{wang2014knowledge} and ComplEx~\cite{complex}. For each model (i.e., text-only + graph model combination), we experimented with both embedding approaches and selected the one that performed best on the validation sets.
All GCNs were configured as follows: 
two edge types (one for edges in ConceptNet and one for the self-loops); 
300 dimensions for each embedding across all layers;
one convolutional layer; 
one additional linear layer (after the convolution); and
ReLU for all activations.
These parameters yielded best average accuracy on the validation sets, so that we chose them uniformly for all models for consistency across our approaches.

The Personalized PageRank threshold $\theta$ for filtering the subgraphs was also tuned as a hyperparameter. We experimented with $\theta$ values of 0.2, 0.4, 0.6, and 0.8. We did not experiment with whole one-hop graphs ($\theta$ = 0.0), as they have been shown to increase in size very rapidly over single hops in ConceptNet ~\cite{wang2019improving}. 

Training (of the combined models) consisted of 140 epochs with a patience of 20 epochs. Batch size and learning rate over all the experiments remained 64 and 0.0001 to make the models comparable to each other. 

\subsection{Results}

Table~\ref{tab:text_graph_models_results} gives an overview of our results. They demonstrate that KES, and thus external knowledge, has the biggest impact on the BreakingNLI test set. 
The accuracy of text-only models is improved, for BERT-based match-LSTM model by 18 percentage points, match-LSTM by 13 percentage points, HBMP by 3 percentage points, and DecompAttn by 8 percentage points. Notably, the most dramatic impact of KES is on the BERT-based match-LSTM model, which is generally the strongest text-only model on the other datasets.   

Despite their competitive performance on SNLI, all text-only models perform significantly worse on the BreakingNLI test set when compared to the SNLI test set, which is consistent with observations from the original BreakingNLI paper. The DecompAttn text-only model shows the biggest drop in performance (28 percentage points) between SNLI and BreakingNLI.  The match-LSTM text-only model shows the smallest drop in performance between SNLI and Breaking NLI -- still a substantial 18 percentage points. In contrast, KES shows only modest decreases in performance between SNLI and BreakingNLI when a GloVe- or BERT-based match-LSTM text model is used, with accuracy decreasing only 5 and 8 percentage points respectively. However, there is a significant decrease in performance between SNLI and BreakingNLI when KES uses HBMP or DecompAttn as its text model (20 and 26 percentage points respectively), suggesting a potentially complex interaction between text and external knowledge features. Overall, while KES models perform comparably to its text-based counterparts for SNLI, SciTail, and MultiNLI, they perform significantly better on BreakingNLI dataset. 

These results support three important claims. First, they demonstrate that KES is modular in that it can be combined with existing text models with different architectures. Second, the KES approach effectively infuses external knowledge into existing entailment models to improve performance on the challenging BreakingNLI dataset. Third, KES is robust to dataset changes that dramatically decrease the performance of other NLI models.

\subsubsection{Comparison to Other KG-based Models.}
Table~\ref{tab:text_graph_models_results} also contains the results for the graph-based models KIM and ConSeqNet. 
Both show comparable performance to match-LSTM KES, with KIM performing best on all datasets.  We discuss important differences between KES, KIM, and ConSeqNet below.

KIM 
introduces an external knowledge-based co-attention mechanism, using five manually engineered features from WordNet for every term pair of words in premise and hypothesis. These features are specific to WordNet relations, which means that the model can only be used with WordNet or comparable KGs with the same set of relations. One can argue that, because KIM depends on WordNet, it is especially suited to BreakingNLI, as WordNet contains exactly the type of lexical information that is targeted by BreakingNLI. 
Another difference between KIM and KES is that it is not clear how to adapt KIM's five engineered features to a different textual entailment system. In contrast, KES is not tied to any particular KG, KG vocabulary, or existing entailment system. One of the practical goals of KES is to develop an approach that is easily adaptable to different datasets, knowledge graphs, and existing entailment models. Although tuning only the PPR threshold as the hyper parameter, our knowledge augmented approaches perform almost on par with KIM on on SNLI and MultiNLI except   BreakingNLI dataset (-4.4 percentage). 

ConSeqNet, similar to our model and unlike KIM, does provide an architecture to plug in any text based entailment model. However, there are two primary differences between our work and ConSeqNet.  First, we are the first to encode the graph structure of the knowledge graph where as ConSeqNet uses on the concepts mentioned in text encoding them using RNNs. Also, in comparison to ConSeqNet, our approach performs better with different entailment models over all the datasets. Particularly, on the BreakingNLI dataset, our implementation of ConSeqNet shows a drop in performance in comparison to its text-based method. This is in turn surprising and may need further investigations.  

In summary, in addition to the performance goals of KIM and ConSeqNet, KES seeks to infuse entailment models with knowledge in a way that is modular and sensitive to graph structure, independent of a specific KG.

\subsubsection{Harnessing External Knowledge.}
 Table~\ref{tab:ppr_graph_analysis} shows the average number of concepts (nodes) and relations (edges) in contextual subgraphs generated by KES, ConSeqNet, and KIM, excluding those that were explicitly mentioned in the premise and hypothesis texts.  Unlike ConSeqNet and KIM, KES is able to use a great amount of external knowledge that is related to the premise and hypothesis but not explicitly mentioned. As observed in prior work~\cite{wang2019improving}, expanding subgraphs by even one hop results in very large graphs, making PPR filtering very important. 

\begin{table*}
\centering
\resizebox{0.75\linewidth}{!}{
\begin{tabular}{|l|c|c|c|c|c|c|}  
\hline
\multirow{2}{*}{\textbf{PPR}} & \multicolumn{2}{|c|}{\textbf{Scitail (17.74*)}} & \multicolumn{2}{|c|}{\textbf{SNLI (11.5*)}} &
\multicolumn{2}{|c|}{\textbf{MultiNLI (17.5*)}}\\
&  
\textbf{Edges} & 
\textbf{Nodes} &  
\textbf{Edges} & 
\textbf{Nodes}&  
\textbf{Edges} & 
\textbf{Nodes}\\
\hline
0.2 & 42.65	& 10.14 & 80.29 & 19.83 & 76.27 & 16.15\\
0.4 & 26.72 & 7.48 & 25.70 & 8.15 & 33.82 & 6.48  \\
0.6  & 15.53 & 4.35 & 14.08 & 4.65 & 23.97 & 3.44 \\
0.8  & 11.67 & 3.04 & 9.98 & 3.18 & 20.27& 2.05 \\
\hline
\multirow{2}{*}{ConSeqNet} & 0 & 0 (17.74*) & 0 & 0 (11.5*) &0 &0 (17.5*)\\
& \multicolumn{6}{|l|}{
No edges or new concepts are added from ConceptNet.}\\
\hline
KIM & \multicolumn{6}{|l|}{
Features based on fixed WordNet relations. No new concepts are added.}\\
\hline
\end{tabular}
}
\caption{
Average number of nodes and edges (not explicitly mentioned in text) in combined premise and hypothesis subgraphs by PPR threshold.
 *Average number of concepts explicitly mentioned in each premise and hypothesis text.
}
\label{tab:ppr_graph_analysis}
\end{table*}

\section{Discussion}
\noindent
\textbf{Negative Results}: In Table~\ref{tab:text_graph_models_results}, we observe two results that did not confirm our hypotheses: (1) the reduced text+graph improvement on BreakingNLI for HBMP, and (2) lower text+graph performance for DecompAttn on SciTail ($>$ 2 percentage points). We are investigating these issues, but one possible explanation for the reduced improvement on HBMP is that it is one of the few text based models that has a large final hidden layer (14K feature vector) in comparison to the features from the GCN model (900) which is possibly biasing the final classifier towards the text-based features.

\smallskip
\noindent
\textbf{Personalized PageRank Threshold: } Our initial plan for using PPR thresholds was to make it a preprocessing step and fix one threshold for a dataset on a base model. However, as shown in Table~\ref{tab:text_graph_models_results}, using PPR thresholding as a hyperparameter for each model trained showed better performance. Also, the PPR threshold, in particular $0.8$ filters very few concepts that aren't mentioned in premise and hypothesis text, whereas contextual subgraphs from $0.2$  can contain the equal number of concepts from external knowledge as mentioned in text (Table~\ref{tab:ppr_graph_analysis}). PPR filtering is just one possible method for reducing noise that results from neighborhood-based expansion techniques. In our future work, we intend to investigate a different filtering approach where only those paths that connect premise and hypothesis are included.

\smallskip
\noindent
\textbf{Dataset characteristics. }We evaluated our KES approach on NLI datasets that are widely used in the literature. However, there has been criticism regarding the way these datasets are created and the resulting biases that can be exploited by learning algorithms \cite{glockner2018breaking,li2019several,gururangan2018annotation,poliak2018hypothesis}. Even in our work, in Table~\ref{tab:text_graph_models_results} where we see that the DecompAttn model is consistently improved by KES on SNLI, MNLI, and BreakingNLI, we also see the opposite effect on SciTail. Some qualitative analysis of the SciTail dataset showed us that use of KG can negatively impact the performance because of high overlap between premise and hypothesis terms. 

Text-based models trained on SNLI perform significantly worse on the BreaklingNLI test set, consistent with the results reported above. Notably, the estimated human performance on the BreakingNLI test set is higher than that of the original SNLI test set, providing further evidence that models that perform well on SNLI but poorly on BreakingNLI are poor approximations for human inference. On the other hand, NLI models that generalize well to BreakingNLI are more likely to be better approximations for human-like inference. The complexity of the BreakingNLI test set and its characteristics make it the most interesting evaluation set.

\smallskip
\noindent
\textbf{Complexity of Knowledge Graphs and their usage:} As mentioned above, the current state-of-the-art for BreakingNLI is the KIM model, which achieves an 83\% accuracy, while our best performing KES model (KES with the match-LSTM text model) achieves an accuracy of 79\%.  This difference can be attributed to aspects of the KIM model that make it particularly well suited to the BreakingNLI dataset at the expense of model flexibility and generality.  KIM relies on WordNet, which has lexical information that aligns very closely with the challenging aspects of the BreakingNLI. This focus clearly benefits performance on the task. However, WordNet is relatively small (117k triples, i.e., edges) compared to ConceptNet (3.15M triples) and has a very specific scope that is unlikely to cover the broad classes of entailment that occur in natural language. For example, recognizing textual entailment may depend on world knowledge that is not lexical in nature.  In such cases it would be necessary to invoke a model that is not primarily focused on lexical knowledge.  This is one of the motivations behind the KES approach: to support very large KGs (e.g., ConceptNet) and to avoid dependencies on any single KG or domain area. An important topic for future work will be to understand the shortcomings of various knowledge sources, how to manage choosing the appropriate knowledge sources for a given task, and to continue exploring graph filtering and selection methods to leverage large scale KGs while minimizing noise. KIM mitigates the noise issue by using a restricted set of relations to provide greater focus and minimize intrusion of potentially irrelevant knowledge. Again, this is a characteristic of KIM that will not necessarily generalize well to other NLI datasets, such as SciTail, which may depend less on hyper- and hyponym relations, and more on knowledge about everyday physical objects and processes.

\section{Conclusion}

In this paper, we presented a systematic approach for infusing external knowledge into the textual entailment task using contextually relevant subgraphs extracted from a KG and 
encoded with graph convolutional networks. These graph representations are combined with standard text-based representations into a KG-augmented entailment system 
which yields significant improvement on the challenging BreakingNLI dataset. Additionally, the KES approach is modular, can be used with any knowledge graph, and is generalizable to multiple datasets. In our future work, we plan to consider other KGs and to investigate alternative graph representations. 
Furthermore, it would be interesting to see how KES performs on the popular question answering datasets.

{\small
\bibliography{AAAI-KapanipathiP.9423}

\begin{thebibliography}{}

\bibitem[\protect\citeauthoryear{Auer \bgroup et al\mbox.\egroup
  }{2007}]{auer2007dbpedia}
Auer, S.; Bizer, C.; Kobilarov, G.; Lehmann, J.; Cyganiak, R.; and Ives, Z.
\newblock 2007.
\newblock Dbpedia: A nucleus for a web of open data.
\newblock In {\em The semantic web}. Springer.
\newblock  722--735.

\bibitem[\protect\citeauthoryear{Bowman \bgroup et al\mbox.\egroup
  }{2015}]{bowman2015large}
Bowman, S.~R.; Angeli, G.; Potts, C.; and Manning, C.~D.
\newblock 2015.
\newblock A large annotated corpus for learning natural language inference.
\newblock In {\em Proceedings of Empirical Methods in Natural Language
  Processing Conference},  632--642.

\bibitem[\protect\citeauthoryear{Chen \bgroup et al\mbox.\egroup
  }{2018}]{chen2018neural}
Chen, Q.; Zhu, X.; Ling, Z.-H.; Inkpen, D.; and Wei, S.
\newblock 2018.
\newblock Neural natural language inference models enhanced with external
  knowledge.
\newblock In {\em Proceedings of the 56th Annual Meeting of the Association for
  Computational Linguistics (Volume 1: Long Papers)},  2406--2417.

\bibitem[\protect\citeauthoryear{Devlin \bgroup et al\mbox.\egroup
  }{2019}]{devlin2019bert}
Devlin, J.; Chang, M.-W.; Lee, K.; and Toutanova, K.
\newblock 2019.
\newblock Bert: Pre-training of deep bidirectional transformers for language
  understanding.
\newblock In {\em Proceedings of the North American Chapter of the Association
  for Computational Linguistics Conference: Human Language Technologies, Volume
  1 (Long and Short Papers)},  4171--4186.

\bibitem[\protect\citeauthoryear{Glockner, Shwartz, and
  Goldberg}{2018}]{glockner2018breaking}
Glockner, M.; Shwartz, V.; and Goldberg, Y.
\newblock 2018.
\newblock Breaking nli systems with sentences that require simple lexical
  inferences.
\newblock In {\em Proceedings of the 56th Annual Meeting of the Association for
  Computational Linguistics (Volume 2: Short Papers)},  650--655.

\bibitem[\protect\citeauthoryear{Gururangan \bgroup et al\mbox.\egroup
  }{2018}]{gururangan2018annotation}
Gururangan, S.; Swayamdipta, S.; Levy, O.; Schwartz, R.; Bowman, S.; and Smith,
  N.~A.
\newblock 2018.
\newblock Annotation artifacts in natural language inference data.
\newblock In {\em Proceedings of the North American Chapter of the Association
  for Computational Linguistics Conference: Human Language Technologies, Volume
  2 (Short Papers)},  107--112.

\bibitem[\protect\citeauthoryear{Harabagiu and
  Hickl}{2006}]{harabagiu2006methods}
Harabagiu, S., and Hickl, A.
\newblock 2006.
\newblock Methods for using textual entailment in open-domain question
  answering.
\newblock In {\em Proceedings of the 21st International Conference on
  Computational Linguistics and the 44th annual meeting of the Association for
  Computational Linguistics},  905--912.
\newblock Association for Computational Linguistics.

\bibitem[\protect\citeauthoryear{Huang \bgroup et al\mbox.\egroup
  }{2019}]{huang2019knowledge}
Huang, X.; Zhang, J.; Li, D.; and Li, P.
\newblock 2019.
\newblock Knowledge graph embedding based question answering.
\newblock In {\em Proceedings of the Twelfth ACM International Conference on
  Web Search and Data Mining},  105--113.
\newblock ACM.

\bibitem[\protect\citeauthoryear{Jeh and Widom}{2003}]{jeh2003scaling}
Jeh, G., and Widom, J.
\newblock 2003.
\newblock Scaling personalized web search.
\newblock In {\em Proceedings of the 12th International Conference on World
  Wide Web},  271--279.
\newblock ACM.

\bibitem[\protect\citeauthoryear{Khot, Sabharwal, and
  Clark}{2018}]{khot2018scitail}
Khot, T.; Sabharwal, A.; and Clark, P.
\newblock 2018.
\newblock Scitail: A textual entailment dataset from science question
  answering.
\newblock In {\em Proceedings of the AAAI Conference on Artificial
  Intelligence}.

\bibitem[\protect\citeauthoryear{Kipf and Welling}{2017}]{kipf2017gcn}
Kipf, T.~N., and Welling, M.
\newblock 2017.
\newblock Semi-supervised classification with graph convolutional networks.
\newblock In {\em International Conference on Learning Representations (ICLR)}.

\bibitem[\protect\citeauthoryear{Lalithsena \bgroup et al\mbox.\egroup
  }{2017}]{lalithsena2017domain}
Lalithsena, S.; Perera, S.; Kapanipathi, P.; and Sheth, A.
\newblock 2017.
\newblock Domain-specific hierarchical subgraph extraction: A recommendation
  use case.
\newblock In {\em International Conference on Big Data (Big Data)},  666--675.
\newblock IEEE.

\bibitem[\protect\citeauthoryear{Li \bgroup et al\mbox.\egroup
  }{2019}]{li2019several}
Li, T.; Zhu, X.; Liu, Q.; Chen, Q.; Chen, Z.; and Wei, S.
\newblock 2019.
\newblock Several experiments on investigating pretraining and
  knowledge-enhanced models for natural language inference.
\newblock {\em arXiv preprint arXiv:1904.12104}.

\bibitem[\protect\citeauthoryear{Liu \bgroup et al\mbox.\egroup
  }{2019a}]{liu2019multi}
Liu, X.; He, P.; Chen, W.; and Gao, J.
\newblock 2019a.
\newblock Multi-task deep neural networks for natural language understanding.
\newblock In {\em Proceedings of the 57th Annual Meeting of the Association for
  Computational Linguistics},  4487--4496.
\newblock Association for Computational Linguistics.

\bibitem[\protect\citeauthoryear{Liu \bgroup et al\mbox.\egroup
  }{2019b}]{liu2019roberta}
Liu, Y.; Ott, M.; Goyal, N.; Du, J.; Joshi, M.; Chen, D.; Levy, O.; Lewis, M.;
  Zettlemoyer, L.; and Stoyanov, V.
\newblock 2019b.
\newblock Roberta: A robustly optimized bert pretraining approach.
\newblock {\em arXiv preprint arXiv:1907.11692}.

\bibitem[\protect\citeauthoryear{MacCartney and Manning}{2009}]{maccartney2009}
MacCartney, B., and Manning, C.~D.
\newblock 2009.
\newblock {\em Natural language inference}.
\newblock Stanford University Stanford.

\bibitem[\protect\citeauthoryear{Miller}{1995}]{miller1995wordnet}
Miller, G.~A.
\newblock 1995.
\newblock Wordnet: a lexical database for english.
\newblock {\em Communications of the ACM} 38(11):39--41.

\bibitem[\protect\citeauthoryear{Moon \bgroup et al\mbox.\egroup
  }{2019}]{moon2019opendialkg}
Moon, S.; Shah, P.; Kumar, A.; and Subba, R.
\newblock 2019.
\newblock Opendialkg: Explainable conversational reasoning with attention-based
  walks over knowledge graphs.
\newblock In {\em Proceedings of the 57th Annual Meeting of the Association for
  Computational Linguistics},  845--854.

\bibitem[\protect\citeauthoryear{Musa \bgroup et al\mbox.\egroup
  }{2019}]{musa2019answering}
Musa, R.; Wang, X.; Fokoue, A.; Mattei, N.; Chang, M.; Kapanipathi, P.; Makni,
  B.; Talamadupula, K.; and Witbrock, M.
\newblock 2019.
\newblock Answering science exam questions using query rewriting with
  background knowledge.
\newblock {\em Automated Knowledge Base Construction}.

\bibitem[\protect\citeauthoryear{Nguyen and Grishman}{2018}]{nguyen2018graph}
Nguyen, T.~H., and Grishman, R.
\newblock 2018.
\newblock Graph convolutional networks with argument-aware pooling for event
  detection.
\newblock In {\em Proceedings of the AAAI Conference on Artificial
  Intelligence}.

\bibitem[\protect\citeauthoryear{Page \bgroup et al\mbox.\egroup
  }{1999}]{page1999pagerank}
Page, L.; Brin, S.; Motwani, R.; and Winograd, T.
\newblock 1999.
\newblock The {PageRank} citation ranking: Bringing order to the web.
\newblock Technical report, Stanford InfoLab.

\bibitem[\protect\citeauthoryear{Parikh \bgroup et al\mbox.\egroup
  }{2016}]{parikh2016decomposable}
Parikh, A.; T{\"a}ckstr{\"o}m, O.; Das, D.; and Uszkoreit, J.
\newblock 2016.
\newblock A decomposable attention model for natural language inference.
\newblock In {\em Proceedings of Empirical Methods in Natural Language
  Processing Conference},  2249--2255.

\bibitem[\protect\citeauthoryear{Poliak \bgroup et al\mbox.\egroup
  }{2018}]{poliak2018hypothesis}
Poliak, A.; Naradowsky, J.; Haldar, A.; Rudinger, R.; and Durme, B.~V.
\newblock 2018.
\newblock Hypothesis only baselines in natural language inference.
\newblock In {\em Proceedings of the North American Chapter of the Association
  for Computational Linguistics Conference: Human Language Technologies},
  180--191.
\newblock Association for Computational Linguistics.

\bibitem[\protect\citeauthoryear{Schlichtkrull \bgroup et al\mbox.\egroup
  }{2018}]{schlichtkrull2018modeling}
Schlichtkrull, M.; Kipf, T.~N.; Bloem, P.; Van Den~Berg, R.; Titov, I.; and
  Welling, M.
\newblock 2018.
\newblock Modeling relational data with graph convolutional networks.
\newblock In {\em Proceedings of the European Semantic Web Conference},
  593--607.
\newblock Springer.

\bibitem[\protect\citeauthoryear{Speer, Chin, and
  Havasi}{2017}]{speer2017conceptnet}
Speer, R.; Chin, J.; and Havasi, C.
\newblock 2017.
\newblock Conceptnet 5.5: An open multilingual graph of general knowledge.
\newblock In {\em Proceedings of the AAAI Conference on Artificial
  Intelligence}.

\bibitem[\protect\citeauthoryear{Talman, Yli-Jyr{\"a}, and
  Tiedemann}{2019}]{talman2019sentence}
Talman, A.; Yli-Jyr{\"a}, A.; and Tiedemann, J.
\newblock 2019.
\newblock Sentence embeddings in nli with iterative refinement encoders.
\newblock {\em Natural Language Engineering} 25(4):467--482.

\bibitem[\protect\citeauthoryear{Trouillon \bgroup et al\mbox.\egroup
  }{2016}]{complex}
Trouillon, T.; Welbl, J.; Riedel, S.; Gaussier, {\'E}.; and Bouchard, G.
\newblock 2016.
\newblock Complex embeddings for simple link prediction.
\newblock In {\em Proceedings of the International Conference on Machine
  Learning},  2071--2080.

\bibitem[\protect\citeauthoryear{Wang and Jiang}{2016a}]{wang2016learning}
Wang, S., and Jiang, J.
\newblock 2016a.
\newblock Learning natural language inference with {LSTM}.
\newblock In {\em Proceedings of the North American Chapter of the Association
  for Computational Linguistics Conference: Human Language Technologies},
  1442--1451.

\bibitem[\protect\citeauthoryear{Wang and Jiang}{2016b}]{wang2016machine}
Wang, S., and Jiang, J.
\newblock 2016b.
\newblock Machine comprehension using match-lstm and answer pointer.
\newblock {\em arXiv preprint arXiv:1608.07905}.

\bibitem[\protect\citeauthoryear{Wang \bgroup et al\mbox.\egroup
  }{2014}]{wang2014knowledge}
Wang, Z.; Zhang, J.; Feng, J.; and Chen, Z.
\newblock 2014.
\newblock Knowledge graph embedding by translating on hyperplanes.
\newblock In {\em Proceedings of the AAAI Conference on Artificial
  Intelligence}.

\bibitem[\protect\citeauthoryear{Wang \bgroup et al\mbox.\egroup
  }{2019}]{wang2019improving}
Wang, X.; Kapanipathi, P.; Musa, R.; Yu, M.; Talamadupula, K.; Abdelaziz, I.;
  Chang, M.; Fokoue, A.; Makni, B.; Mattei, N.; and Witbrock, M.
\newblock 2019.
\newblock {Improving Natural Language Inference Using External Knowledge in the
  Science Questions Domain}.
\newblock {\em Proceedings of the AAAI Conference on Artificial Intelligence}.

\bibitem[\protect\citeauthoryear{Williams, Nangia, and
  Bowman}{2018}]{williams2018broad}
Williams, A.; Nangia, N.; and Bowman, S.
\newblock 2018.
\newblock A broad-coverage challenge corpus for sentence understanding through
  inference.
\newblock In {\em Proceedings of the North American Chapter of the Association
  for Computational Linguistics Conference: Human Language Technologies, Volume
  1},  1112--1122.

\bibitem[\protect\citeauthoryear{Xu \bgroup et al\mbox.\egroup
  }{2019}]{xu2019gnnpower}
Xu, K.; Hu, W.; Leskovec, J.; and Jegelka, S.
\newblock 2019.
\newblock How powerful are graph neural networks?
\newblock In {\em Proceedings of the International Conference on Machine
  Learning}.

\bibitem[\protect\citeauthoryear{Yang \bgroup et al\mbox.\egroup
  }{2019}]{yang2019enhancing}
Yang, X.; Zhu, X.; Zhao, H.; Zhang, Q.; and Feng, Y.
\newblock 2019.
\newblock Enhancing unsupervised pretraining with external knowledge for
  natural language inference.
\newblock In {\em Proceedings of the Canadian Conference on Artificial
  Intelligence},  413--419.

\bibitem[\protect\citeauthoryear{Zhang \bgroup et al\mbox.\egroup
  }{2018}]{zhang2018know}
Zhang, Z.; Wu, Y.; Li, Z.; He, S.; Zhao, H.; Zhou, X.; and Zhou, X.
\newblock 2018.
\newblock I know what you want: Semantic learning for text comprehension.
\newblock {\em arXiv preprint arXiv:1809.02794}.

\end{thebibliography}
\bibliographystyle{aaai}
}

\end{document}